# Context-Aware Saliency Detection for Image Retargeting Using Convolutional Neural Networks

Mahdi Ahmadi [1], Nader Karimi [1], Shadrokh Samavi [1,2]

**Abstract** Image retargeting is the task of making images capable of being displayed on screens with different sizes. This work should be done so that high-level visual information and low-level features such as texture remain as intact as possible to the human visual system, while the output image may have different dimensions. Thus, simple methods such as scaling and cropping are not adequate for this purpose. In recent years, researchers have tried to improve the existing retargeting methods, and they have introduced new ones. However, a specific method cannot be utilized to retarget all types of images. In other words, different images require different retargeting methods. Image retargeting has a close relationship to image saliency detection, which is relatively a new image processing task. Earlier saliency detection methods were based on local and global but low-level image information. These methods are called bottom-up processes. On the other hand, newer approaches are top-down and mixed methods that consider the high level and semantic knowledge of the image too. In this paper, we introduce the proposed methods in both saliency detection and retargeting. For the saliency detection, the use of image context and semantic segmentation are examined, and a novel mixed bottom-up, and top-down saliency detection method is introduced. After saliency detection, a modified version of an existing retargeting technique is utilized for retargeting the images. The results suggest that the proposed image retargeting pipeline has excellent performance compared to other tested methods. Also, the subjective evaluations on the Pascal dataset can be used as a retargeting quality assessment dataset for further research.

## 1 Introduction

In recent years, by the rapid development of multimedia devices, the range of screen aspect ratios is even wider than before. Hence, there is a need for images and videos to be displayed on different devices well. The operation of changing image dimensions and aspect ratio so that the resulting image

Shadrokh Samavi
E-mail: samavi@mcmaster.ca

Mahdi Ahmadi
E-mail: mahdi.ahmadi@ec.iut.ac.ir

Nader Karimi
E-mail: nader.karimi@cc.iut.ac.ir

1 Department of Electrical and Computer Engineering, Isfahan University of Technology, 84156-83111, Iran.
2 Department of Electrical and Computer Engineering, McMaster University, Hamilton, L8S 4L8 Canada.

is compatible with the new device is called image retargeting. In image retargeting, the produced image should contain the essential content and include the minimum distortion.

While there are a lot of methods for image retargeting, no one can be used for all the images, and each has its pros and cons. Moreover, the retargeted images of different ways for a given image do not score similarly by human raters, and selecting between the methods is highly subjective. However, the research is toward finding algorithms that satisfy most of the people.

Image retargeting shares its techniques with some other image processing tasks such as inpainting [20], object removal [27], content amplification [1], and image synthesis [15]. Also, it utilizes different knowledge areas such as graph theory and optimization. There have been various extensions of image retargeting in recent years. For example, stereo image and video processing [26] is the use of image retargeting in a couple of stereo images so that the resulting couple not only has the most relevant content but also there is the least distortion in their disparity map.

There has been a bunch of research on the retargeted image quality assessment (RIQA). In this area, the task is to evaluate the retargeted images so that the evaluation (objective) scores are similar to human (subjective) scores. For this task, there are some datasets, such as RetargetMe [23] and CUHK [18]. These datasets contain the human scores for retargeted images of some retargeting methods, and the RIQA algorithms should produce similar scores for them. The datasets are also used as benchmarks in papers when a new retargeting method is introduced.

Image retargeting has a close relationship with saliency detection. Saliency detection tries to simulate human gaze behavior. The result of saliency detection is a saliency map in which the most critical regions are the ones that most attracts the human gaze. According to the research in saliency detection, the human visual system (HVS) has a mixed top-down bottom-up approach in the human gaze behavior. So, the methods are in top-down, bottom-up and mixed approaches. The top-down processes use semantic and high-level information to produce saliency map, while bottom-up methods use low-level information such as color, contrast, and texture.

Saliency detection can be used as a preprocessing in other applications such as image watermarking [8], image compression [9], object detection [34], and image classification [11]. As in image retargeting, the retargeted image should preserve the most essential content; hence, saliency detection plays an important role. Thus saliency detection is used as preprocessing in most new retargeting methods. Saliency detection methods use different approaches to render this task. Some techniques use the statistical characteristics of a dataset [21], some use cognitive sciences as the real saliency models that should be followed [12], and some use machine learning that similar to statistical models estimate the dataset characteristics [14].

In recent years artificial neural networks, specifically convolutional neural networks (CNN) have gained a lot of attention to them for their supreme abilities in a large variety of image processing tasks. They are used in the automated car driving, medical image processing, object detection and segmentation, and a lot more. From a more statistical viewpoint, CNNs can produce new samples for existing data, estimate a distribution, and distinguish between several distributions. Semantic segmentation, in one of their unique applications, tries to classify images in the pixel level. CNNs can also be used directly to the objectives of this paper, i.e., image retargeting [3], [30], and saliency detection [14], [4].

In this paper, a saliency detection method based on image context and semantics is introduced. The semantic segmentation and context detection tasks are done together and by a shared CNN. The use of image context is based on the observation that different objects have different relative importance in different contexts. For example, a building has a different saliency when it is in a jungle than when it is in urban scenery. In this paper, the objects are segmented by the semantic segmentation network and are assigned their saliency values according to the image context that is another output of the network. Moreover, the low-level image information such as image color and contrast are used in producing the saliency map. Thus, the proposed method in saliency detection can be considered a mixed top-down, bottom-up saliency detection method.

After the saliency map is generated, the image is retargeted by a method inspired by the method presented in [37]. In the proposed method, the image pixels are assigned scaling factors according to their saliency. These scaling factors allow the pixels to move just in two directions. For example, if the retargeting is to reduce the image size in the horizontal direction, and the pixels will move to right and left directions. Finally uniform sampling is performed to the resulting image to produce the retargeted image. The overall diagram of the proposed method in context detection, saliency detection and image retargeting can be seen in Fig. 1.

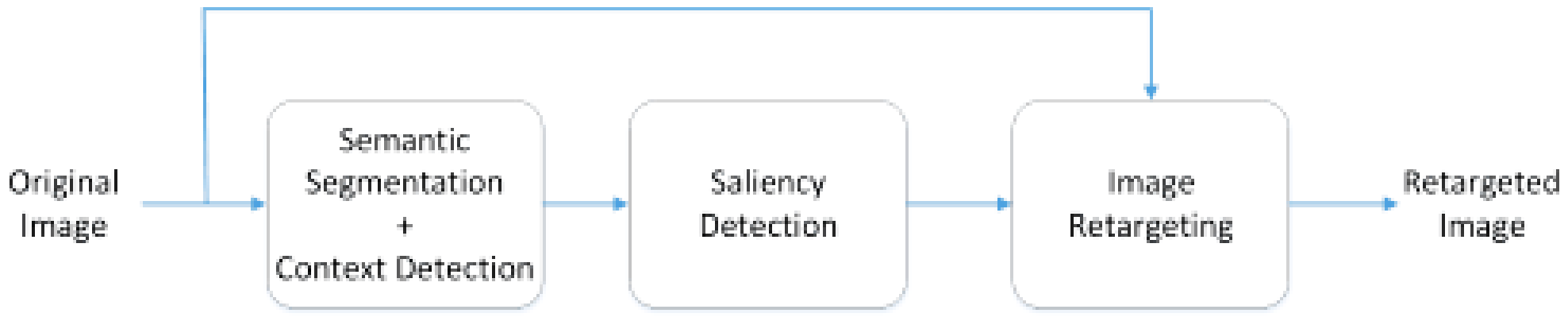

Fig. 1: Overall block diagram of the proposed method

In the next section, we review some existing papers in both saliency detection and image retargeting. Then the proposed methods in both saliency detection and retargeting are introduced in section 3. In section 4, the results of the saliency detection will be compared to some other methods. Also, the retargeted images are compared to some other retargeting purposes. Finally, we will conclude the work in section 5 and discuss possible future work for the proposed research.

## 2 Related Work

As in the proposed image retargeting method is based on saliency detection, in this section, both saliency detection and image retargeting papers are reviewed.

### 2.1 A Review of Saliency Detection Methods

In the saliency detection, the goal is to model the human gaze regions as a map. The older methods were bottom-up and the saliency map was built by low-level image features. One of the primary processes, which was inspired by the human visual system, is the work of Itti et al. in [12]. In this method, image features like color and edge directions are processed in different scales and then the results are combined to build the saliency map. In the newer research, the top-down methods and mixed methods are considered too. For example, the authors in [31], the low-level information is a local binary pattern (LBP) for texture and color contrast both in local and global senses. However, the image focus is the high-level information from which the more important image parts are detected.

In the method presented in [10], the authors solve the saliency detection problem by a graph model and an information-theoretic viewpoint. The authors in [32], make an energy function by image saliency and coherency and then optimize it. The optimization problem is then solved to find the saliency of pixels. In [13], the image color is presented in a high-dimension space, and the extracted color features build the saliency map. In [21], the statistical saliency features are derived from a dataset, then the saliency map for each image is estimated by a Dirichlet random process. Their work shows that some colors and the central image regions are more prone to be salient.

The authors in [14], use a CNN to extract image features in a hierarchical scheme, and the image is examined in multiple scales. Another work that utilizes machine learning tools is the work of [4] in which the authors use LSTM network layers. By using these layers, the neighboring image patches affect each other, and the output is more accurate. The authors in [16], combine the features extracted by a CNN with low-level features such as contrast and color. In the algorithm introduced in [41], the image is fed to two different networks to extract local and global features, and the saliency map is the combination of extracted features.

The saliency map can be obtained by semantic segmentation. For example, authors in [33] make an importance-list for objects and assign different saliency values to different objects that are seg-mented by semantic segmentation. The other example is [22] that produces a saliency map by image geometry, semantic priority, and gradients. In both papers, the generated saliency map is used in image retargeting.

### 2.2 A Review of Image Retargeting Methods

The image retargeting methods are usually classified into two major categories: discrete and continuous methods. In the discrete category, the resulting image is directly constructed by the pixels or patches of the original image. On the other hand, the continuous methods mainly solve this problem by a non-uniform sampling on the original image.

One of the first examples of the discrete category is the work by [1] in which vertical or horizontal seams are selected iteratively to be deleted. The method is called seam carving and is the basis of many other algorithms. For example, the authors in [37], first find the seams to be deleted. Then the pixels on individual seams are considered as groups that have different scaling factors. Finally, the retargeted image is constructed by fusing the pixels given their scaling factors.

Another important discrete method is an image cropping. The image in this method is so cropped that the resulting image contains the most essential content from either attention or aesthetic viewpoints [5]. One of the first attention-based methods is the work by [29] in which a face detection algorithm is used to find the most important image regions. In [5], the authors build a dataset that shows the most aesthetically essential image regions by bounding boxes. They train a CNN that estimates the boxes. Finally the retargeted image is produced by cropping the obtained boxes.

Some discrete methods define the retargeted image as a combination of the patches in the original image. For example, in the shift-map approach [20] the relative shift of individual patches is modeled by a graph, and the task is to optimize a function on this graph. Another criterion is bidirectional similarity, which is optimized to build the retargeted image [27].

In the continuous category, the retargeted image is built by a non-uniform sampling on the original image. This is in contrast with uniform sampling that is also called scaling. In the scaling method, the image content is not considered, and the change of aspect ratio in important image regions makes them seem distorted. The method introduced in [35] is an example of continuous methods. In this method, the image is scaled by local scaling factors so that the image contents in the important regions, which are identified by the significance map, remain undistorted. In another work that is done by [36], a system of equations is solved to obtain the new coordinations of each pixel in a video stream. The authors in [19], developed their method which is called axis aligned (AA). In AA, two energy terms one for rigidity and one for similarity are defined. Finding the new coordinates for the pixels is then to easily solve a quadratic optimization problem. The authors in[3], use two pre-trained VGG16 [28] networks as parts of an end to end retargeting network. The network produces a saliency map by self-supervised learning that is the basis of non-uniform sampling of the image.

There are some methods that utilize a combination of discrete or continuous methods for image retargeting. These so-called multi-operator methods were firstly introduced by [25] that the three operations of scaling, seam carving and cropping were applied iteratively to the image. They found the best mixture of operations by modeling the problem as a dynamic programming problem. In another example, the authors in [7] compare the retargeted image to the original image in every iteration to find the best operator. In every iteration, the operator can be one of seam carving, scaling, cropping, and warping operators and the comparison criteria is perceptual similarity measure.

## 3 The Proposed Method

As seen in Fig. 1, the proposed image retargeting pipeline consists of context detection, semantic segmentation, saliency detection, and finally retargeting operation. For each image, the saliency map is first generated based on image context and semantic segmentation. Afterward, the image and its saliency map are fed into an image retargeting method. In this section, the saliency detection and then

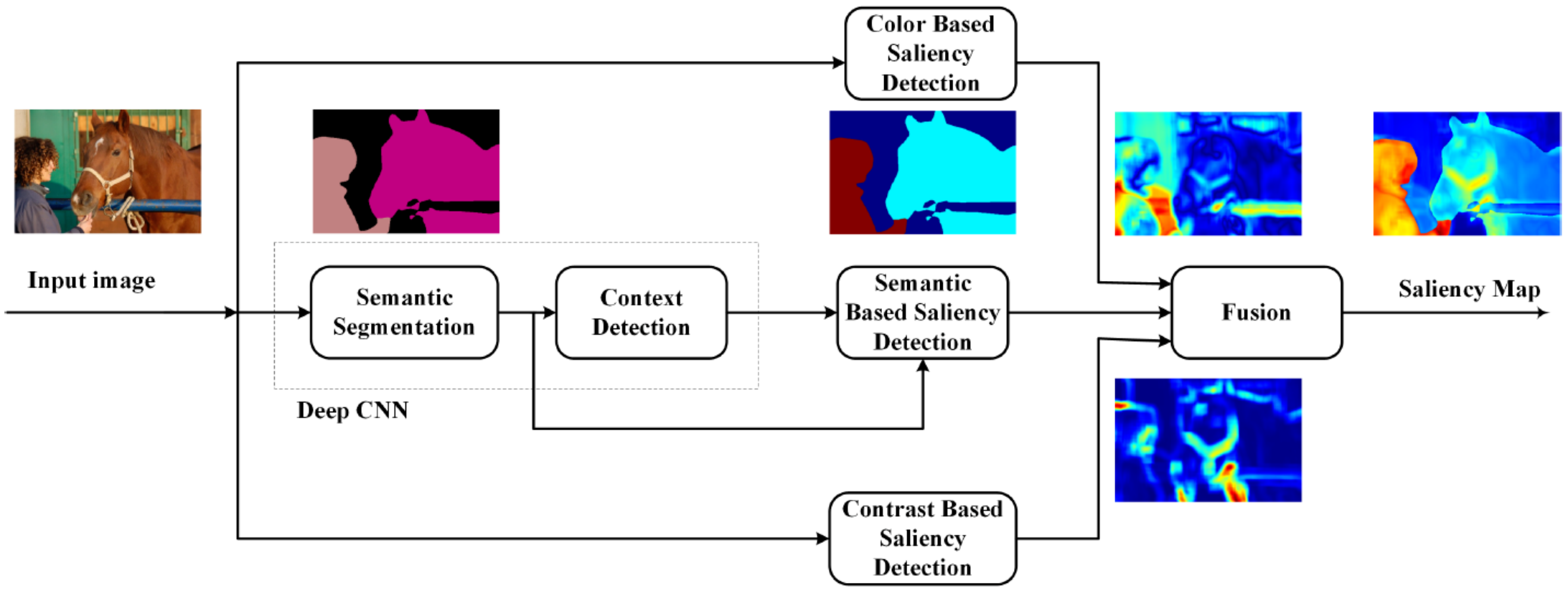

Fig. 2: Overall diagram of the proposed method in saliency detection

parts of the saliency detection, and so are presented in that section.

## 3.1 Saliency Detection

In the proposed saliency detection method, a combined top-down, bottom-up method is introduced. The bottom-up elements of the proposed method are made of the color contrast and amplitude contrast of the image pixels. The first is roughly named the color part, and the second one is called the contrast part of the saliency map. On the other hand, the top-down element is constructed by image semantic segmentation and image context. In this paper, the image context has a determining role in the usage of semantic data. The overall saliency detection method is shown in Fig. 2.

According to this Figure, the final saliency map is a linear combination of color-based, contrast-based, and semantic segmentation based saliency maps and is calculated by:

$$S_{fused} = \gamma_1 S_{SSEG} + \gamma_2 S_{CN} + \gamma_3 S_{CL}, \quad \gamma_1 + \gamma_2 + \gamma_3 = 1 \tag{1}$$

In this equation, $S_{fused}$ is the combined saliency map and $S_{SSEG}$, $S_{CN}$, and $S_{CL}$ are its semantic segmentation part, contrast part, and color part respectively. The parameters $\gamma_1$, $\gamma_2$, and $\gamma_3$ are three constant values. In the following parts of this section, each of the blocks of Fig. 2 will be explained in detail.

### *3.1.1 The Semantic Part of the Saliency Map*

To build the semantic part of the proposed saliency detection method, image semantic segmentation and image context are required. Semantic segmentation is produced by a convolutional neural network named PSP-Net (Pyramid Scene Parsing) [40]. The image context is also obtained by a neural network. These two networks can be considered as a unified network which is illustrated in Fig. 3. As seen in this image, the semantic segmentation network can be split into two parts named encoder and decoder. The encoder, also known as feature extractor, extracts features that are useful for both decoder and context detection.

*Semantic Segmentation* The utilized semantic segmentation network in this paper is PSP-Net. We used a trained version of this network on the Pascal dataset [6] that its weights are available at [39].

This network is one of the state-of-the-art networks for semantic segmentation. In this network, due to the pyramid structure of the feature extraction layers, the image features are combined in different scales.

The semantic segmentation methods are evaluated by mIOU (mean Intersection Over Union) which is calculated by:

$$IoU = \frac{TP}{TP + FP + FN} \tag{2}$$

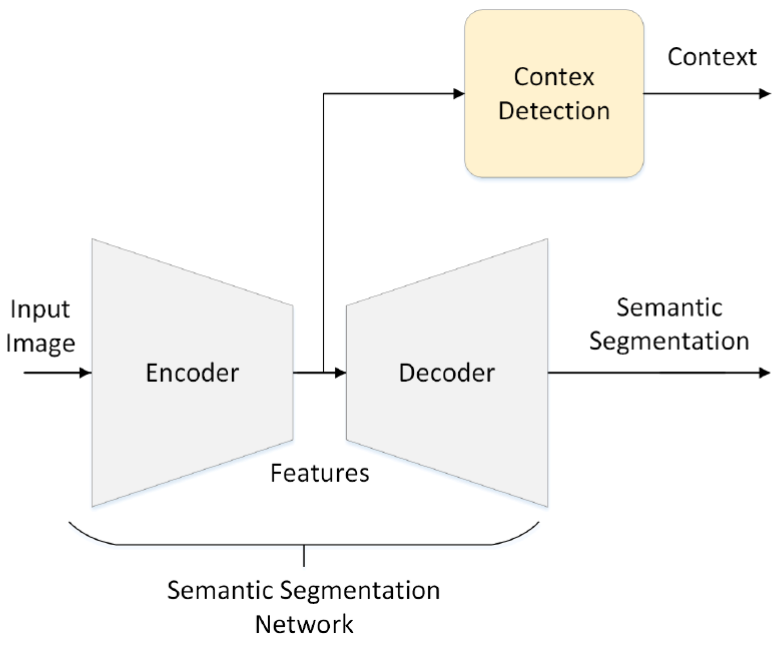

Fig. 3: The simplified structure for semantic segmentation and context detection.

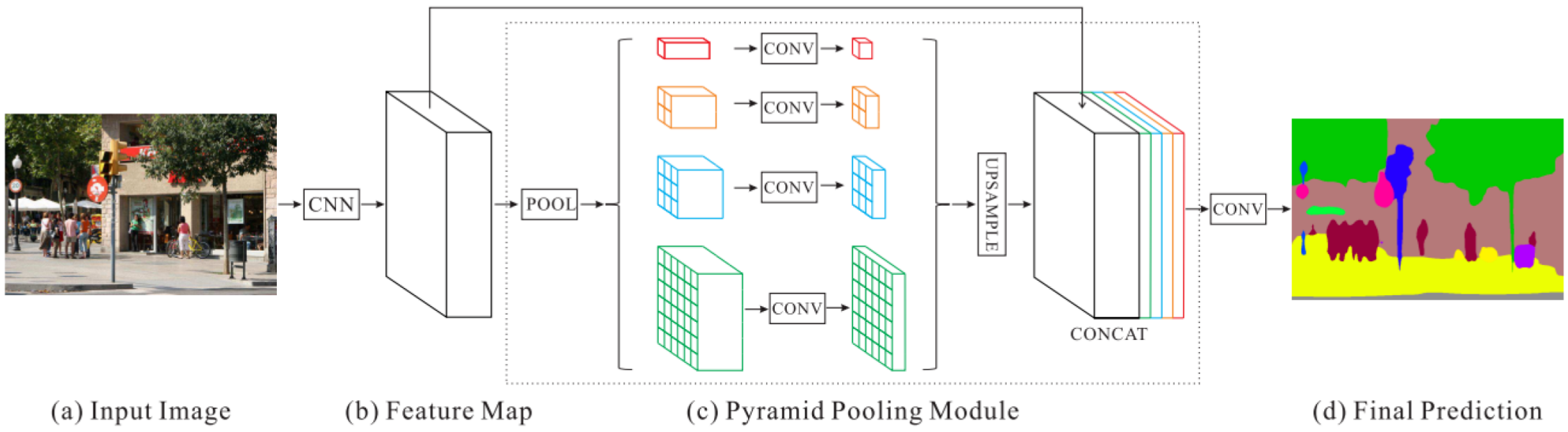

Fig. 4: The structure of the utilized semantic segmentation network (PSP-Net [40])

$$mIoU = \frac{1}{N_C} \sum_{k=1}^{N_C} IoU_k \tag{3}$$

PSP-Net has the mIOU of 82.6 on the Pascal dataset that makes it one of the best available semantic segmentation networks.

*Context Detection* In this paper, all images are categorized into 5 contexts. The proposed contexts are “pet”, “other animals”, “Vehicle”, “indoor” and “other”. For example, the images in the “indoor” context may contain a bedroom, kitchen, or other indoor stuff. The reason to separate the contexts “pet” and “other animals” is to keep balance in class numbers for training. Also, it gives us more freedom in ordering the objects that are part of the proposed semantic-based saliency detection. There is no specific context for humans because humans can exist in images of other contexts and according to the context has their own priority. In this work, all images in the Pascal dataset were categorized into the above-mentioned contexts by annotators.
To detect the image context, we utilized the features extracted by the decoder part of the semantic segmentation network. For PSP-Net, the decoder is considered the end of the pyramid pooling module as seen in Fig. 4. The context detection network should be able to detect the context of images without the familiar objects in the Pascal dataset.
To handle this problem, 200 augmentation images are added to the Pascal training set from the COCO dataset [17]. The augmentation images are selected so that they contain none of the familiar objects in the Pascal dataset. As will be seen in experimental results, this dataset augmentation increases the accuracy of the context detection network.
The feature extractor part of the PSP-Net produces a tensor of the depth 4096 feature channels. Each slice of this tensor is a matrix representing the existence of one of the 4096 features in different image regions. The proposed context detection network works with the average value in each channel. This concept is known as the global average pooling and represents the average value of each feature in the whole image. The output of this stage is a vector containing 4096 elements. This vector then is fed to a soft-max with five neurons, and the neuron with the maximum value shows the image context. The proposed context detection network is shown in Fig. 5.

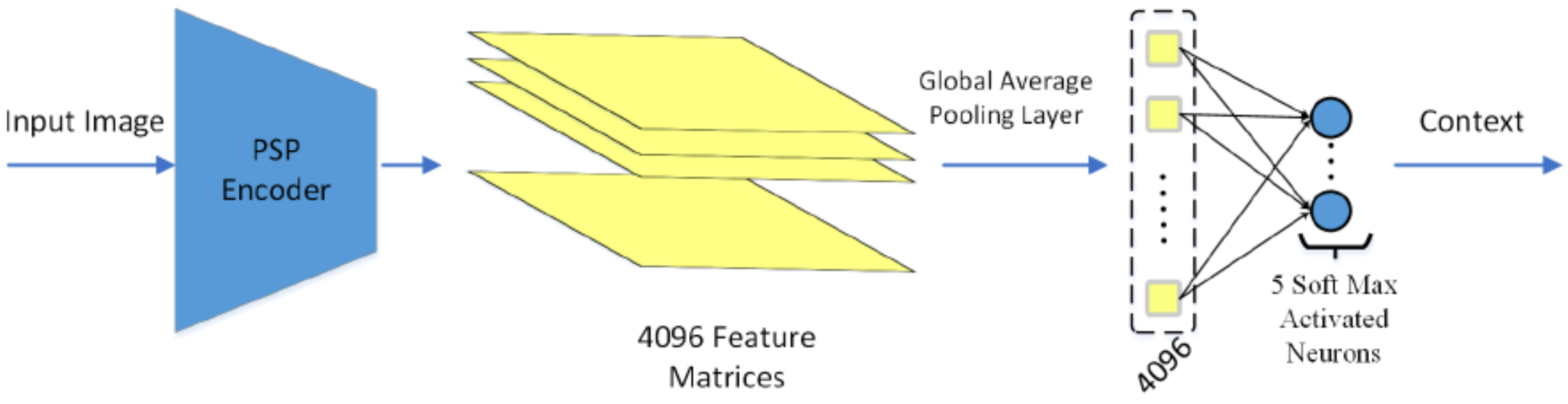

Fig. 5: The proposed structure for context detection.

*Using Semantics and Context* To produce the semantic part of the saliency map, pixels are ordered according to their class and the image context. The relative priorities of objects are stored in a number of look-up-tables (LUT). As the number of image contexts is 5, there are 5 previously filled LUTs each of which stores the relative importance of pixel classes in one of the contexts. There is another LUT to which the user can store a custom priority according to a specific task. The diagram of the pixel-order mechanism is demonstrated in Fig. 6. As seen in this figure, a signal from the context detection network determines a suitable order.

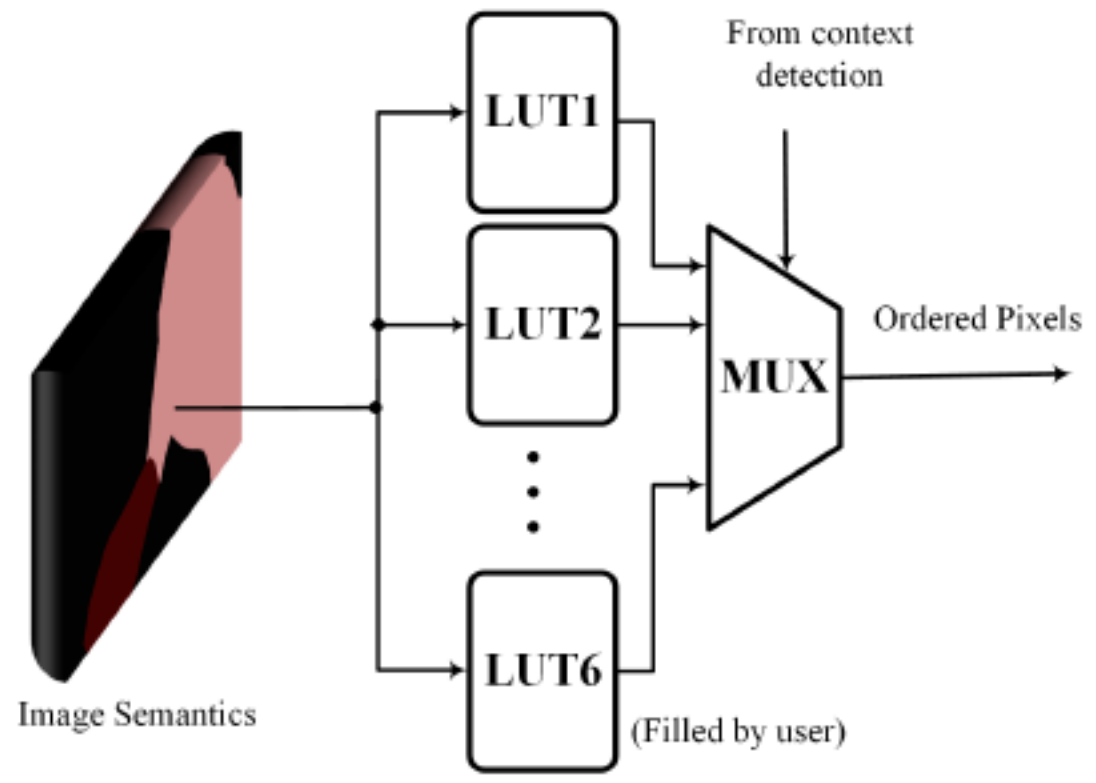

Fig. 6: Ordering the pixels based on semantic segmentation. In this diagram, some already filled look-up tables (LUT) store the object priorities. For each pixel, the corresponding LUT to its context, determines its priority.

The pixel priorities in different contexts are presented in Table 1. In this table, the most important objects at each row get the priority, and then the second and third priorities are assigned. As seen in this table, all objects have the same priority in the context "other".

Table 1: Relative priority of objects in different contexts.

| **Class** | aeroplane | bicycle | bird | boat | bottle | bus | car | cat | chair | cow | diningtable | dog | horse | motorbike | person | pottedplant | sheep | sofa | train | tvmonitor |
|---|---|---|---|---|---|---|---|---|---|---|---|---|---|---|---|---|---|---|---|---|
| **Pet** | 3 | 3 | 2 | 3 | 3 | 3 | 3 | 1 | 3 | 2 | 3 | 1 | 3 | 3 | 1 | 3 | 2 | 3 | 3 | 3 |
| **Animal** | 3 | 3 | 2 | 3 | 3 | 3 | 3 | 2 | 3 | 2 | 3 | 2 | 2 | 3 | 1 | 3 | 2 | 2 | 3 | 3 |
| **Vehicle** | 1 | 1 | 2 | 1 | 3 | 1 | 1 | 3 | 2 | 3 | 3 | 3 | 3 | 1 | 1 | 3 | 3 | 3 | 1 | 2 |
| **Indoor** | 3 | 3 | 3 | 3 | 2 | 3 | 3 | 3 | 2 | 3 | 2 | 3 | 3 | 3 | 1 | 2 | 3 | 2 | 3 | 2 |
| **Other** | 1 | 1 | 1 | 1 | 1 | 1 | 1 | 1 | 1 | 1 | 1 | 1 | 1 | 1 | 1 | 1 | 1 | 1 | 1 | 1 |

In this stage, the saliency value of image pixels are assigned as follows:

$$S_{SSEG}(i,j) = \exp(-(Ordering(i,j) - 1)) \tag{4}$$

In this equation, $S_{SSEG(i,j)}$ is the semantic segmentation based saliency of $(i,j)$th pixel that is a function of its priority, i.e., $Ordering(i,j)$. The background pixels are set the priority of $\infty$ in this equation.

#### *3.1.2 The Contrast Part of The Saliency Map*

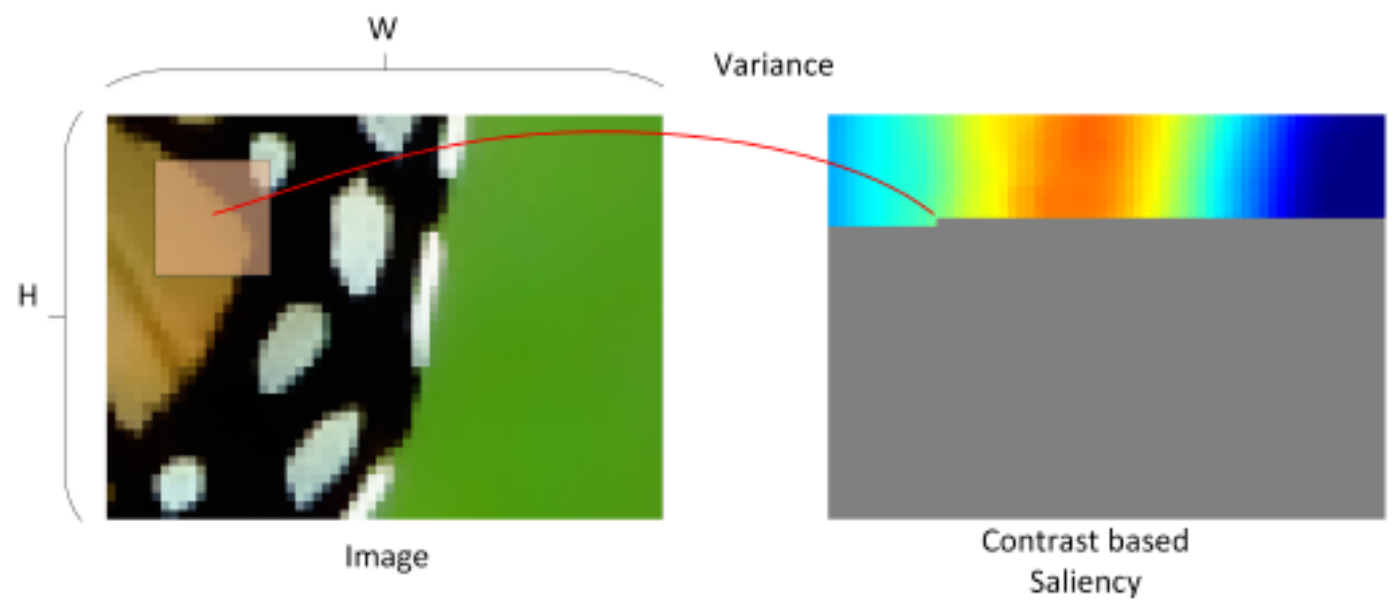

Fig. 7: The diagram of contrast part of the saliency map. The variance of a sliding window, is considered the saliency of its central pixel.

The contrast of different image regions is a good criterion for their saliency. In this part, contrast is defined in the gray-scale values of the image pixels. As seen in Fig. 7, a sliding window moves on the image, and its variance is considered as the contrast of the window. The calculated contrast is assigned to the central pixel of the window which is obtained by:

$$S_{CN} = \frac{1}{N^2}\sum_{i=1}^{N}\sum_{j=1}^{N}(V_{i,j} - \bar{V})^2 \tag{5}$$

in which $V_{i,j}$ is the gray-value of the pixel $(i,j)$, $\bar{V}$ is the mean value of the window, $N$ is its size and $S_{CN}$ is the contrast-based saliency of the window.

#### *3.1.3 The Color Part of The Saliency Map*

The proposed color saliency is in terms of the difference between the local pixel values and a global color value. The assumption to obtain it is that the pixels outside the dominant image colors whose colors are more different than the dominant colors are more salient. The stages of obtaining the color saliency are shown in Fig. 8.

The dominant color is determined by representing the image in the HSV color space. The H channel ($C^H$) and V channel ($C^V$) are separately processed to build the saliency map. First, the H (V) channel histogram of the image in 10 bins in calculated. The center of the dominant bin determines image dominant color $C^H_{Global}$ ($C^V_{Global}$). For every pixel, its local color $C^H_{Local}$ ($C^V_{Local}$) is defined by the average in H (V) channel in its containing window ($\omega$). The containing window is a window that the pixel is located at its center. As the colors in H channel are distributed circularly, the distance (saliency) in this channel is obtained by:

$$C^H_{Local}(i,j) = \frac{1}{N^2}\sum_{(i,j)\in\omega} C^H(i,j) \tag{6}$$

$$d = \left|C^H_{Local}(i,j) - C^H_{Global}\right| \tag{7}$$

$$C^H_{Contrast}(i,j) = \min\{d, 1-d\} \tag{8}$$

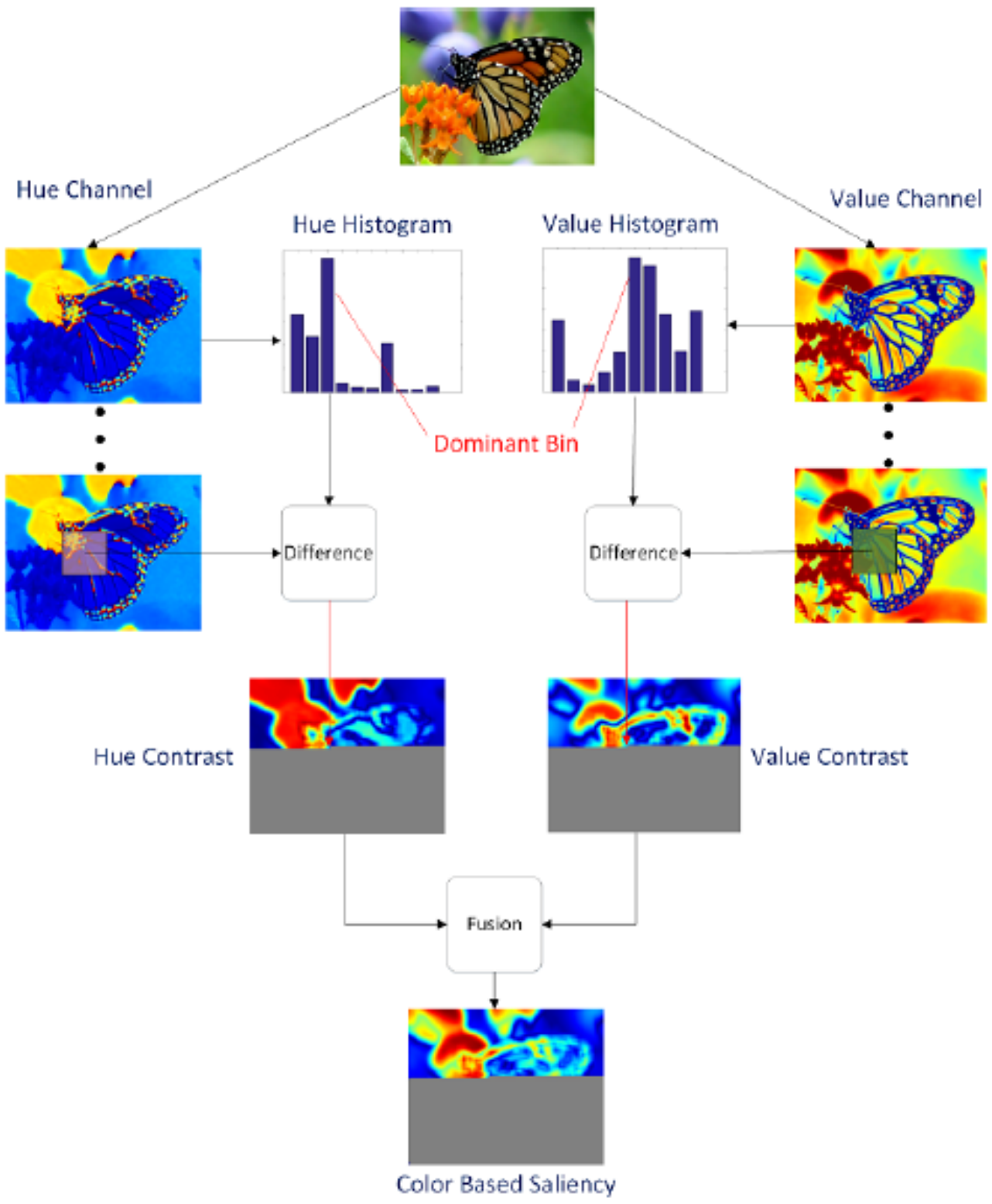

Fig. 8: The diagram of color based saliency detection. As can be seen, the difference of a local value in a window and the global one which is obtained from histogram in each of the H and V channels forms the color contrast. Fusing these two contrasts, creates the color part of the saliency map.

and in V channel, the saliency is calculated by:

$$C^V_{Local}(i,j) = \frac{1}{N^2} \sum_{(i,j)\in\omega} C^V(i,j) \tag{9}$$

$$C^V_{Contrast}(i,j) = \left|C^V_{Local}(i,j) - C^V_{Global}\right| \tag{10}$$

At last, the color saliency in the pixel $(i,j)$, i.e. $S_{CL}(i,j)$ is obtained by:

$$\begin{aligned} S_{CL1}(x,y) &= C^H_{Contrast}(i,j) + C^V_{Contrast}(i,j) \\ S_{CL}(i,j) &= \frac{S_{CL}(i,j)}{\max\{S_{CL1}\}} \end{aligned} \tag{11}$$

In all equations, the window size $N$ is set to 20. The method can be simplified by rendering a simple preprocessing. We can apply a mean filter by the window size of $N$ on the H and V channels to obtain $C^H_{Local}$ and $C^V_{Local}$ for all pixels. This idea simplifies the coding and reduces the processing time. By viewing the proposed color saliency from another point of view, it is easily concluded that the color saliency is the difference between a global value and the mean filtered H and V channels.

### *3.1.4 The Final Saliency Map*

To fuse the different acquired saliency maps, we use equation 1 in which the parameters $\gamma_1$, $\gamma_2$, and $\gamma_3$ are set to 0.25, 0.25, and 0.5 respectively. In some studies on the saliency dataset introduced in [2], the role of pixel location is studied and it is concluded that the central pixels are more prone to be salient. This phenomenon is called center-priority. In the proposed method, the center-priors are utilized to build the final saliency map. So, the final saliency map is obtained by a pixel-wise multiplication of the fused saliency map to a Gaussian window that is shown in the following equation:

$$S_{final}(i,j) = C_1 + \exp\left(\frac{-\left((i-\frac{W}{2})^2 + (j-\frac{H}{2})^2\right)}{C_2.\max\{W,H\}}\right).S_{fused}(i,j) \tag{12}$$

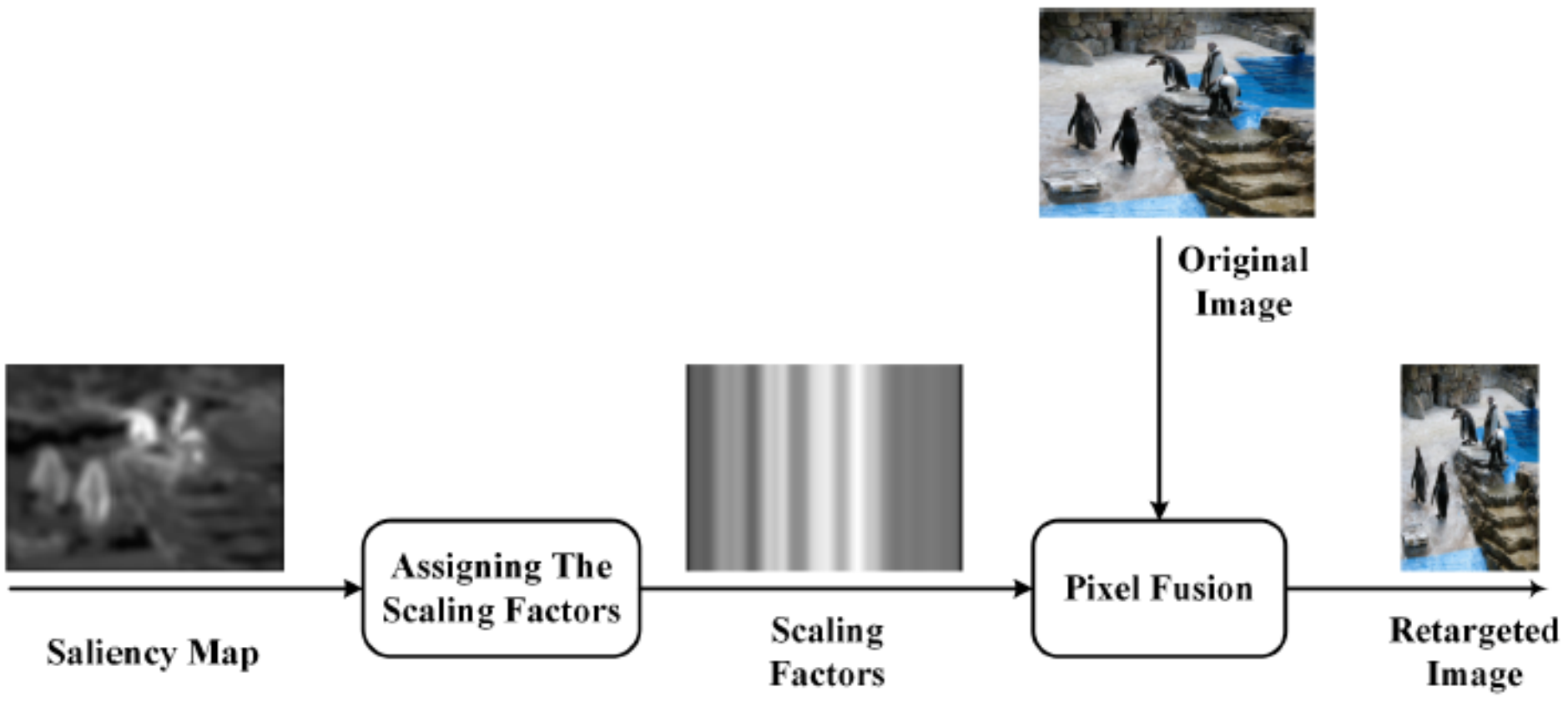

Fig. 9: Overall diagram of the proposed method in the image retargeting

In this equation, $S_{final}(i,j)$ is the saliency value of the pixel $(i,j)$ after applying center-priors, $H$ and $W$ are the image dimensions, and $C_1$ and $C_2$ are two constant values set to 2 and 40 respectively.

## 3.2 Image Retargeting

After obtaining the saliency map, the image can be retargeted by any image retargeting technique. We can consider the image retargeting as an operator that its inputs are the original image and its saliency map. The proposed image retargeting technique is a modified version of the work of [37]. In the proposed method, every pixel is first assigned a scaling factor based on its saliency. Then, the pixels are considered as movable springs. They move based on their lengths (scaling factor) and form an image with the size to which the original image will be retargeted. Finally, uniform sampling is applied to the new image that can be considered a non-uniform sampling on the original image. As vertical image retargeting is similar to the horizontal retargeting, in this section we just discuss the horizontal image retargeting. The overall stages of the image retargeting scheme can be seen in Fig. 9.

### *3.2.1 Assigning Scaling Factors to The Pixels*

Similar to two variable probability density functions, the image saliency map is a two-variable function. In the proposed method, the column-wise sum is calculated for the saliency map (like marginal probability functions). For a $H \times W$, image the result is a vector of size $W$ calculated by:

$$Acc(m) = \sum_{n=1}^{H} S_{final}(m,n) \tag{13}$$

In this equation, $Acc(m)$ is the summation of the $m$th column of saliency map and $S_{final}(m,n)$ is the $(m,n)$th pixel saliency. To have a retargeted image of size $H \times W'$ $(W' < W)$, the scaling factors are calculated by:

$$S_0(m) = W'.\frac{Acc(m)}{\sum_{i=1}^{W} Acc(i)} \tag{14}$$

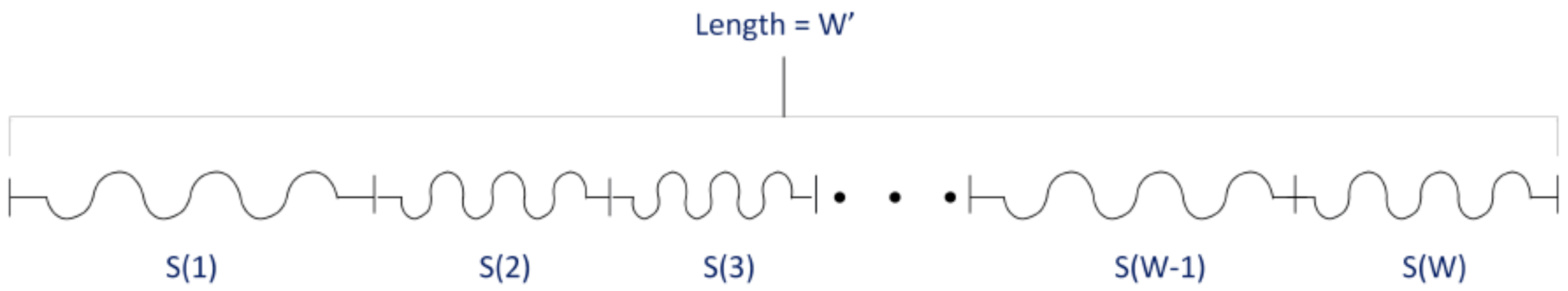

Fig. 10: The springs model for fusing the pixels

Based on this equation, the sum of the scaling factors is the new image width ($W$) because:

$$\sum_{m=1}^{W} S_0(m) = \sum_{m=1}^{W} W'.\frac{Acc(m)}{\sum_{i=1}^{W} Acc(i)} = W' \tag{15}$$

A good condition is that none of the scaling factors become greater than 1. To accomplish this condition, the variable Ng is defined that stands for the number of scaling factors higher than 1. The scaling factors are then updated iteratively by Equation 16 till the condition $N_g = 0$ is yielded.

$$S_t(m) = \frac{(W' - N_g).S_{t-1}(m)}{\sum_{n=1, S_{t-1}(m)<1}^{W} .S_{t-1}(n)}, \quad W' \leq W \tag{16}$$

In this equation, $S_t(m)$ an $S_{t-1}(m)$ are the current and previous scaling factor of the $m$th column. The scaling factors are the same for all pixels of the column. This makes the retargeted image has fewer distortions in the vertical direction.

*3.2.2 Pixel Fusion*

The description for the sampling is called pixel fusion, as called in the paper [37]. In this description, each pixel is modeled by a spring, with a length equal to the pixel's scaling factor. This model is illustrated in Fig. 10. The retargeted image is built by sampling on the springs in equal distances of 1 unit. Due to the equality of the scaling factors on each column of the image, all rows are sampled equally . The sampled value for the $(i', j)$th location on the springs $I_{out}(i', j)$ is a linear combination of its surrounding pixels and is calculated by:

$$I_{out}(i', j) = C_1 I_{in}(m, j) + C_3 I_{in}(m + k, j) + \sum_{n=1}^{k-1} S_{in}(m + n, j) I_{in}(m + 1, j) \tag{17}$$

Which is made of three parts. The two first parts include the $(m; j)$ and $(m + k; j)$ pixels on the original image and the third part includes their $k$ □ 1 middle pixels. $S_{in}$s are the central scaling factors. The symbols and lengths are shown in Fig. 11. It can be inferred from this figure that $C1$ and $C3$ are scaling factors and equation 17 is a linear combination of the image pixels based on their scaling factors. From Fig. 11, it is also evident that the sum of scaling factors to form a pixel of the retargeted image is 1.

## 4 Experimental Results

In this section, the simulation results for both saliency detection and retargeting are presented. Besides, the datasets and learning details of the context detection network are discussed. The comparison to other methods in both saliency detection and retargeting is also done.

The utilized hardware for this paper was PC system with 64GB of RAM and CPU core i7 3.5GHz from Intel©. The system was reinforced by Titan 1080 GTX GPU from NVIDIA®. The operating system was 64 bit Windows 7. The codings were in MATLAB, and Python. Also, the neural networks were implemented on a TensorFlow platform.

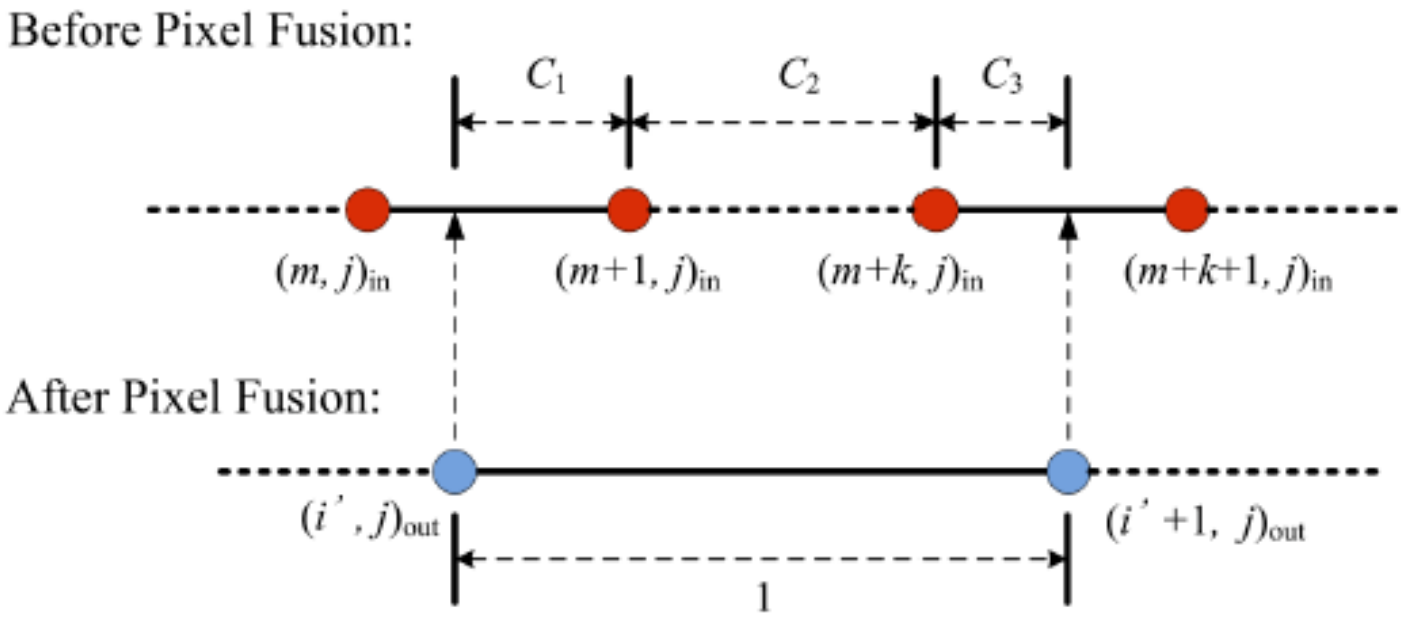

Fig. 11: The description of scaling factors in terms of distances [37]

Table 2: The pascal images in terms of context

| | Pet | Animal | Vehicle | Indoor | Other |
|---|---|---|---|---|---|
| **Train Set** | 200 | 227 | 456 | 190 | 127 |
| **Validation Set** | 43 | 57 | 93 | 46 | 25 |
| **Test Set** | 240 | 297 | 537 | 227 | 148 |

Table 3: The CUHK images summary

| | Pet | Animal | Vehicle | Indoor | Other |
|---|---|---|---|---|---|
| **# of images** | 1 | 5 | 12 | 1 | 38 |

## 4.1 The Datasets

In this paper three datasets are used: Pascal VOC2012 [6], COCO [17], and CUHK [18]. Pascal is a well-known dataset in image classification, object detection, and semantic segmentation. Its semantic segmentation image set contains 1464 training images and 1449 validation images. As its test set is not publicly available, the validation images are used as test set in the literature. The semantic segmentation network is pre-trained on the Pascal training set and we used the trained network. We split the Pascal training set into two parts: first 1200 images are used for the training and 264 images for the validation process. The Pascal dataset was categorized into the five contexts of this paper, and the details can be seen in Table 2. We used a subset of 200 images from the COCO dataset. This subset of the COCO dataset is used to augment the Pascal training set in order to train the context detection network.

The CUHK dataset contains 57 images and is produced by the Chinese University of Hong Kong (CUHK) researchers for Retargeted Images Quality Assessment (RIQA) purposes. We retarget all of the images in this dataset in addition to Pascal images by our proposed retargeting method. We categorized the CUHK images into the 5 proposed contexts and the results are seen in Table 3.

## 4.2 Saliency Detection Results

In this part, first, the accuracy of the context detection network and then the saliency detection results and their comparison to some other saliency detection methods are presented.

### *4.2.1 Context Detection*

The context detection network is made of a global average pooling and five soft-max neurons and gets its input from a semantic segmentation network. In the training phase, the semantic segmentation network was freeze, and its weights were not updated. The only weights that were updated were the soft-max weights. The selected optimizer was RMSprop, and the number of epochs was 400. The batch size was set to 80, and the learning rate was chosen as $10^{-5}$. The function to be optimized was

Table 4: Image context detection accuracies

| | Pascal | CUHK |
|---|---|---|
| Augmented | 0.935 | 0.860 |
| Not Augmented | 0.944 | 0.754 |

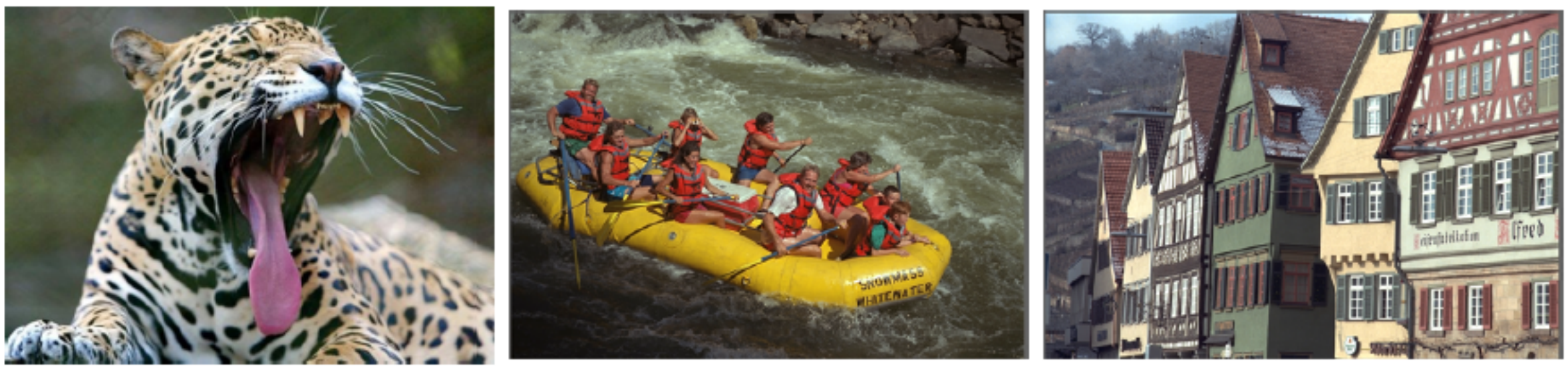

(a) Context is "other animals" but is predicted "pet"

(b) Context is "vehicle" but is predicted "other"

(c) Context is "other" but is predicted "vehicle"

Fig. 12: Some failure examples of the context detection network

accuracy that is calculated by:

$$Accuracy = \frac{\sum_{c=1}^{N_C} TP_C}{n_T} \quad (18)$$

In this equation, $TP_C$ (True Positive) is the number of images in the context $C$, which are detected correctly. $N_C$ is the number of contexts that is 5 in this paper, and $n_T$ is the number of images in batch.

Accuracy is again selected as the criteria for the performance of the context detection network. The accuracy results are presented for the Pascal dataset test set as well as for CUHK dataset. The results are presented for both situations when the Pascal dataset is augmented with COCO images, and when it is not augmented. The results can be seen in Table 4.

As seen in Table 4, the context detection network has better accuracy on the CUHK dataset in the augmented mode. For this reason, the network trained on augmented dataset is selected for saliency detection. However, the accuracy is not 100%, and some images are detected in the wrong context. Some examples of false context detection are presented in Fig. 12.

*4.2.2 Comparison of the Obtained Saliency Maps*

The stages to obtain the proposed saliency map are shown in the Fig. 13. The two first images are from Pascal and the other ones are from the CUHK dataset. In the third image, the snowman is not detected by the semantic segmentation network and thus is considered as the background. The reason is that snowman is not one of the detectable objects for the utilized semantic segmentation network.

The proposed saliency detection and retargeting methods will show their power in the near future when the semantic segmentation networks will have gained better accuracies on more detectable objects.
To display the proposed saliency detection performance, we do a visual comparison to some other methods. In Fig. 14, some images and their saliency maps by the proposed method and other methods are shown. The images are from the two datasets of Pascal and CUHK. As the proposed saliency detection method utilized image segmentation, it has a promising output. For example, in the first row of Fig. 14, the methods GBVS [10] and Itti [12] detect a non-accurate map around the plane.

In the methods HDCT [13] and HSAL [38], some parts of the plane are not detected as salient. The mentioned objection is also valid for most other images. Some different successful results of the proposed method compared to the other methods are shown in the second, fifth and seventh rows.

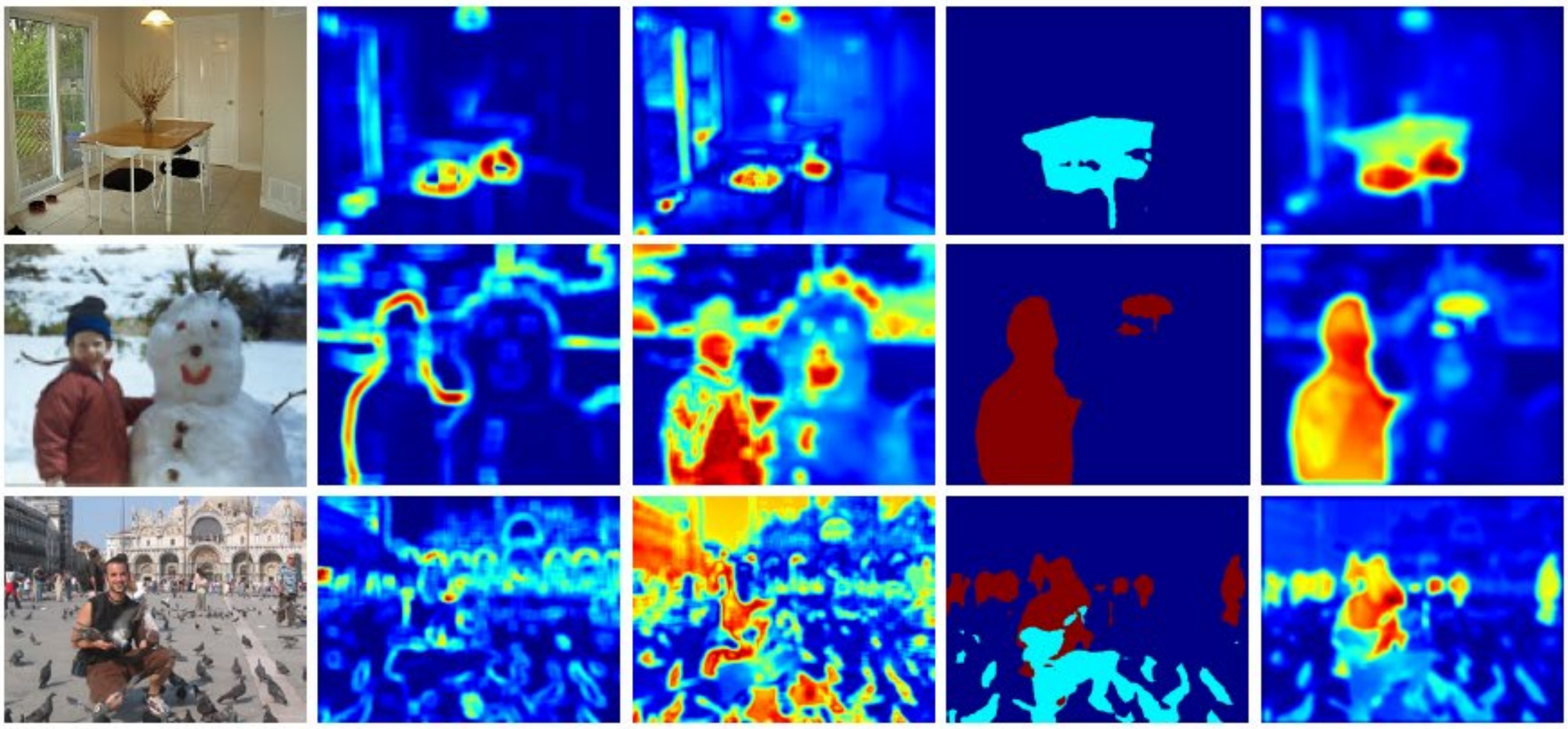

(a) Original image (b) The Contrast part (c) The Color part (d) The Semantic part (e) The Final Map

Fig. 13: Steps to build the saliency map

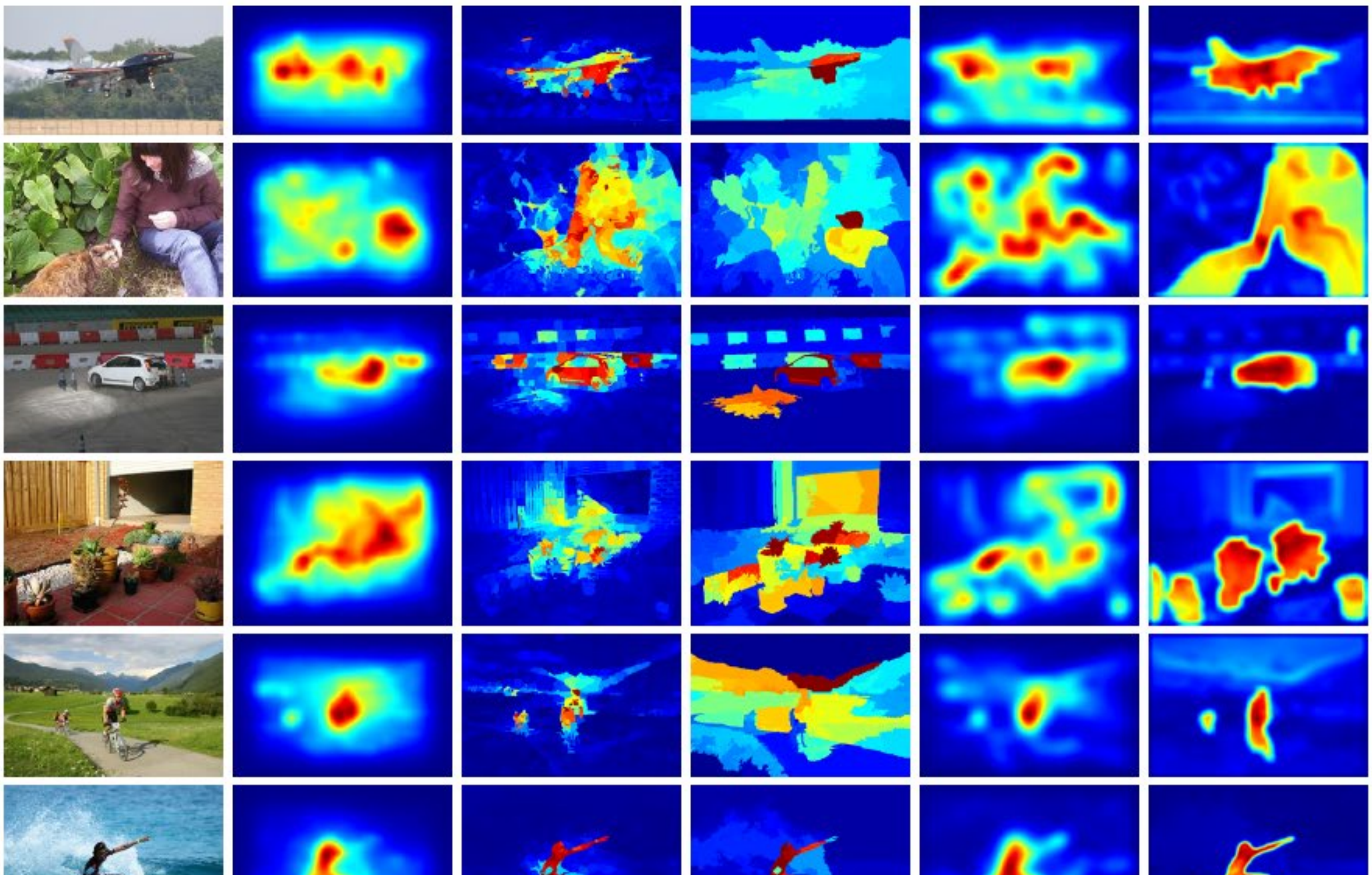

(a) The image (b) GBVS [10] (c) HDCT [13] (d) HSAL [38] (e) Itti [12] (f) Ours

Fig. 14: The comparison between the proposed method and some other methods in saliency detection

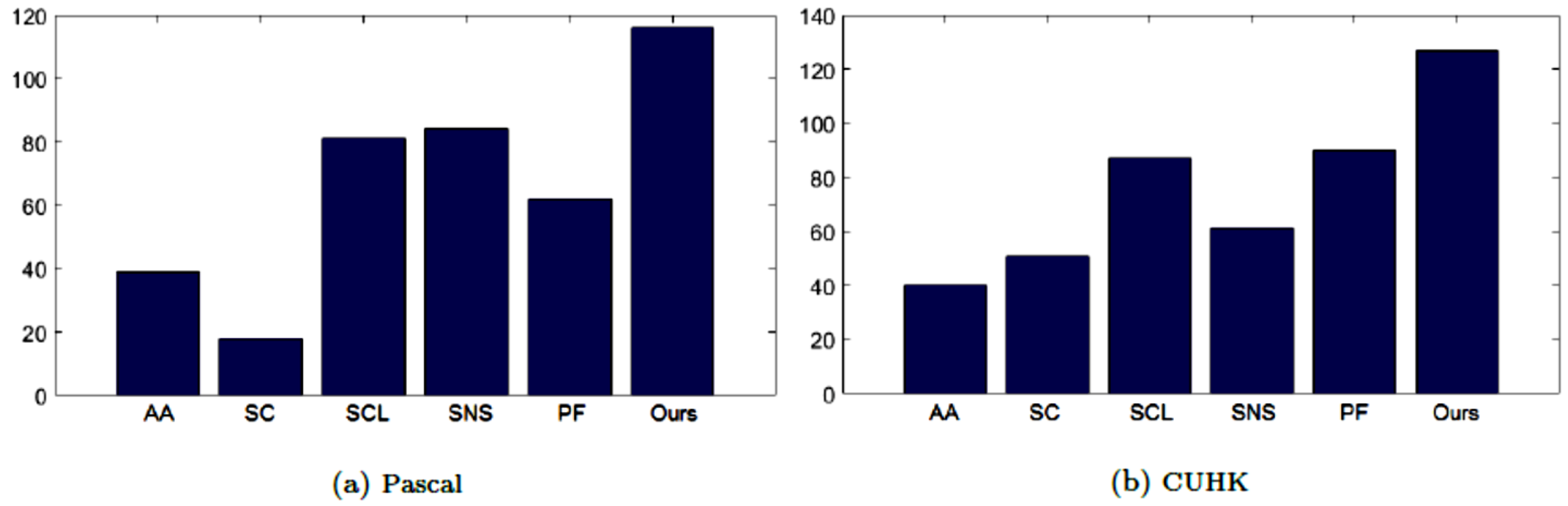

**Fig. 15: The results of the first criterion for subjective evaluations on Pascal and CUHK. The vertical axis is number of the times that the method is selected first.**

## 4.3 Retargeting Results

The retargeted images are evaluated on both the CUHK and Pascal datasets. As subjective evaluations are very time consuming, just 50 images of the Pascal dataset were selected. To collect these 50 images, the first 10 images of each context in test-set were selected. The retargeted images and subjective scores which are represented here specifically the Pascal results can be used as a RIQA dataset for researchers. The other retargeting methods to which the proposed method was compared, are Axis-Aligned (AA) [19], Forward Seam Carving (SC) [24], Scaling (SCL), Scale and Stretch (SNS) [35] and Pixel Fusion (PF) [37].

The scoring is based on image comparisons i.e. every image is retargeted by different methods and 8 volunteers order the retargeted images. The most preferred retargeted image is selected first and so on. For each image, the six retargeted images are shown at the same time to the volunteer but in random order. Also, the original image is available for him/her. To facilitate the ordering task to the volunteers, a MATLAB GUI was built.
The ordered images can be analyzed in different ways. We analyze them in two ways and show that the proposed retargeting method is better than the other compared methods. The two mentioned criteria are discussed as follows.

*First criterion:* For each retargeting method, the number of times that the method is selected first by the evaluators can be considered as a criterion. In this condition, the number is the summation for all evaluators and all dataset images. The results for this criterion are shown in Fig. 15. The diagrams are for both Pascal and CUHK datasets. Based on the diagrams, the proposed method is better than other methods.
These scores can be assigned to the methods when the images are separated into contexts. To do this, the summation is calculated over all images in the context. The results are shown in Fig. 16. According to Fig. 16, the proposed method has the best performance when working with images of the contexts “pet” and “other animals” at the Pascal dataset. As images in CUHK are mostly in the context “other” (see Table. 3), the diagram of CUHK is unbalanced. However, the proposed method has the best scores on the images in this context as well as in the “vehicle” and “pet” contexts.

*Second Criterion:* In some retargeting papers, the comparisons between the retargeted images are two by two. It means that the evaluators must select the superior method between two retargeted images each time. In this case, for 6 different methods, there are $\binom{6}{2} = 15$ comparisons which makes the evaluations very time-consuming. To tackle the problem, some comparisons are skipped in those papers. The proposed ordering method results in a complete two by two comparison, which can be executed faster. For example, when a retargeted image is ranked third in the ordering, it is inferior to the first and second methods and superior to the three remaining methods. From this point of view, the second criterion is defined as the number of times that a method is considered better than another method in two by two comparisons. The number is a summation of all of the images in a dataset and for all evaluators. The results for the second criterion are shown in Fig. 17. In this figure, the vertical axis is the number of times that a method is selected first in all two by two comparisons for all images and all evaluators. Based on the diagrams in Fig. 17, the proposed method is again better than other methods.

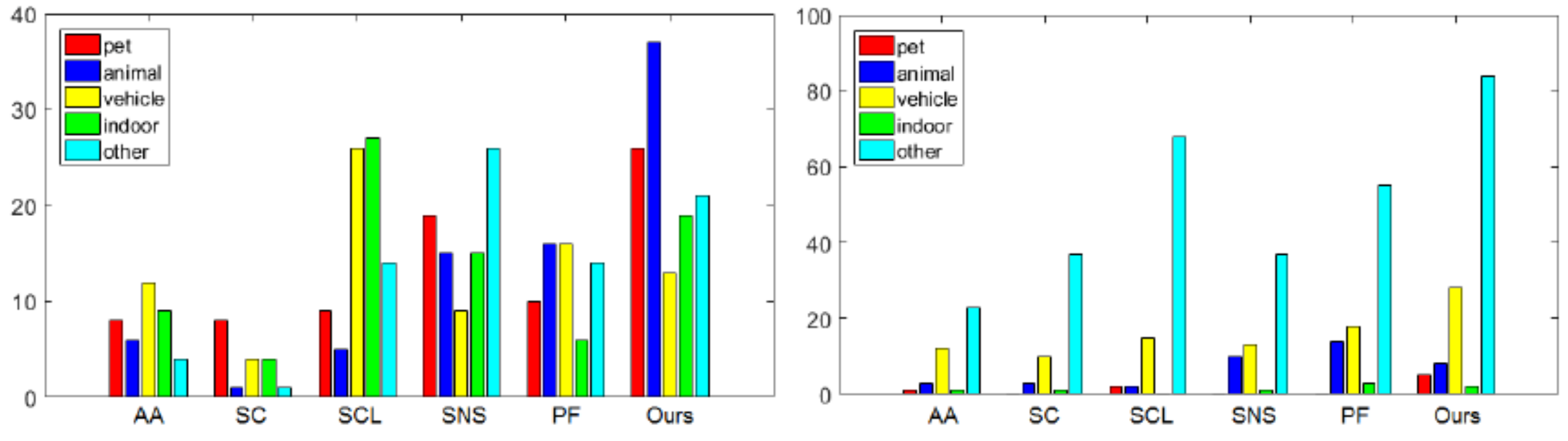

(a) Pascal (b) CUHK

Fig. 16: The results of the first criterion for subjective evaluations on Pascal and CUHK datasets separated by context.

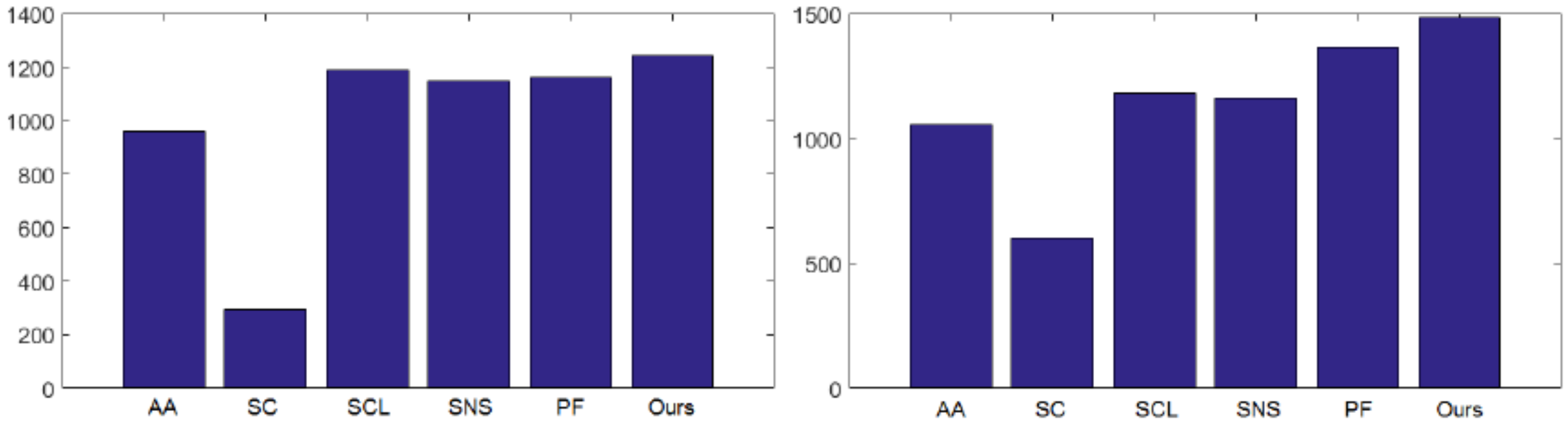

(a) Pascal (b) CUHK

Fig. 17: The results of the second criterion for subjective evaluations on Pascal and CUHK.

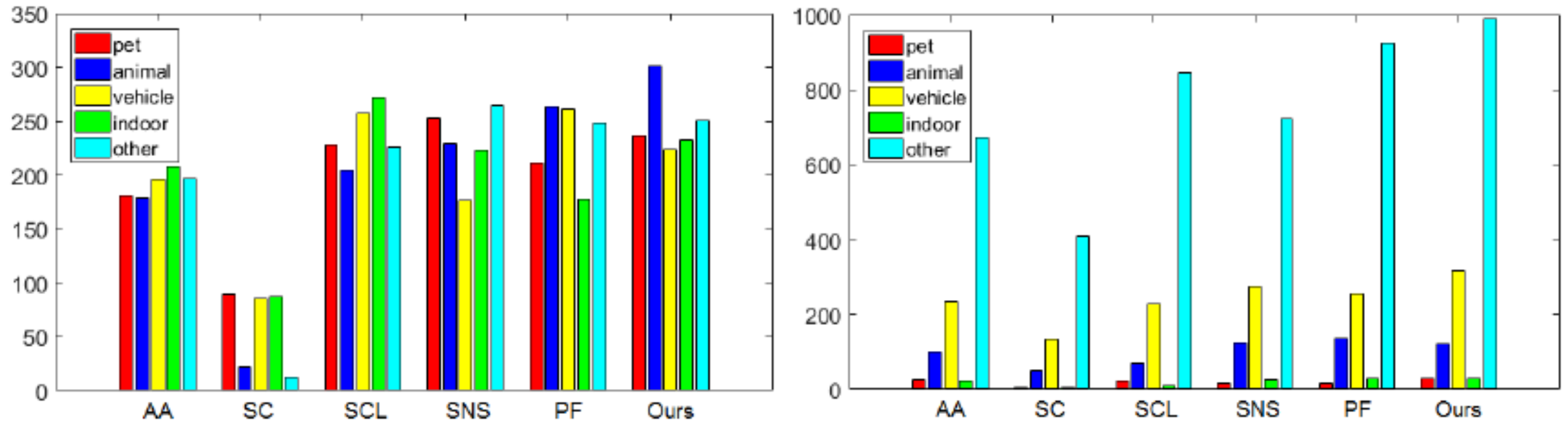

(a) Pascal (b) CUHK

Fig. 18: The results of the second criterion for subjective evaluations on Pascal and CUHK separated by context.

The scores for the second criterion can be calculated within contexts. The results for the context-wise scoring are shown in Fig. 18 for both Pascal and CUHK datasets. According to this figure, the proposed method has the best performance on the “other animals” context of the Pascal dataset. In the CUHK dataset, the best contexts for the proposed method are “other” and “vehicle.”

To obtain the scores of the second criterion, we assigned the score 5 to the first ranked method, 4 to second-ranked and so on. The last ranked method thus has a score of 0. It is obvious that the scores are the number of times that a method wins against the other methods. These scores are then added together on all images and all evaluators to build the second criterion. Therefore, if in the future works we use the ordering method, we can obtain the results for two by two comparisons.

### 4.4 Final Explanations and Failure Examples

In this section, some retargeted images by different methods and then the failure cases of the proposed method are shown and discussed. In Fig. 19, the retargeted images by different methods are shown. As shown in this figure, different retargeting methods have different characteristics and work well on images with specific features. For example, the seam-carving process [24] has acceptable results on images that are full of high-frequency textures like the fourth row of Fig. 19. On the other hand, the methods AA [19] and scaling better preserve the lines in the images (sixth row). However, the scaling method does not consider the image content and scales the whole image equally. The failure examples of the scaling method are the second and the fourth rows. The AA method fails when dealing with images of faces. The SNS method has the best result on the image of the last row, but on the face image in the fifth row fails. In the PF method, the image borders are distorted but the interior content of the images is preserved. The proposed method keeps the most important content of the images. It also keeps the vertical lines of the images intact like two last rows. The proposed method is the best method for the face image in the fifth row.

However, the proposed method like other retargeting methods fails when working with some images. Some failure cases are seen in Fig. 20. The reason for the failures is that the proposed method assigns equal scaling factors to all pixels in columns. For this reason, by sudden changes in the scaling factor from a column to the next column, the straight lines are deflected (the car and TV images). This reason also causes distortion in the less salient columns. For example, in the soccer image one of the player’s feet is on the columns that have less accumulated saliency and thus becomes smaller than the other foot. The other example is the bicycle image that the same occurs to its wheels.

## 5 Conclusions

In this paper, a saliency detection method and a retargeting method were introduced. The proposed saliency detection method is considered as a mixed top-down bottom-up method. The high-level information was image semantic segmentation and image context that a neural network was designed to detect the image context. On the other hand, the low-level information was color contrast and gray-scale contrast. The saliency results showed that the proposed saliency detection method has a good understanding of the image content.

The proposed image retargeting method was a modified version of the Pixel Fusion method presented in [37]. The comparisons to other retargeting methods showed the superiority of the proposed approach to them. To render the comparisons, we used a subjective evaluation of the retargeted images. The results of this subjective evaluation can be used as a dataset for the quality assessment of retargeted images.

For future works, the proposed retargeting method can be generalized to other tasks. In many cases, we need to retarget a streaming video to a screen with different aspect ratio than the original video. For example, when a person watches a video from YouTube, content-aware video retargeting can be a better choice than simple cropping or scaling of the video frames. There is also some research in hardware implementation of specific image retargeting techniques. Therefore, the other possible future work can be the hardware implementation of the retargeting method to achieve real-time results.

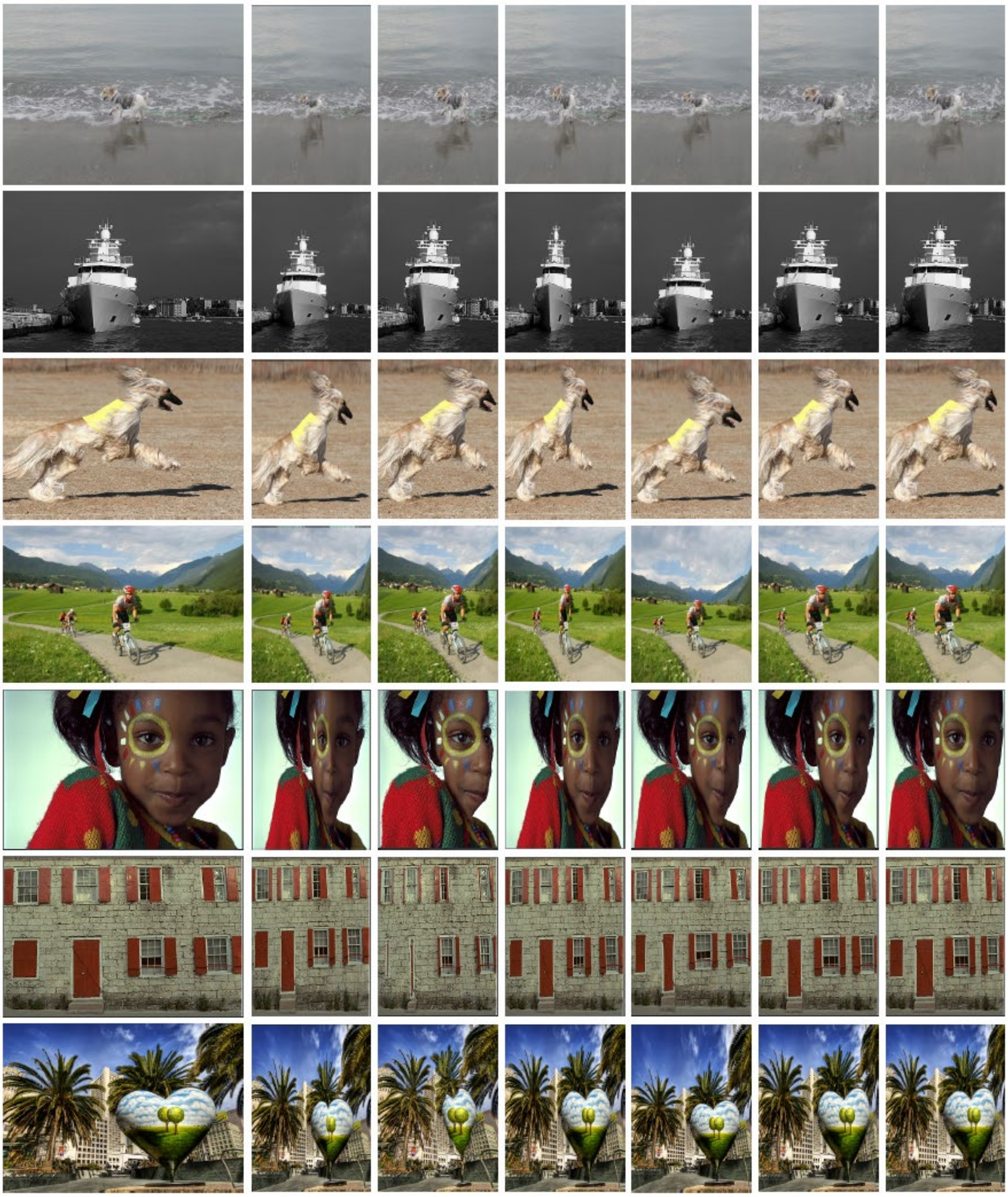

(a) The original image (b) AA[19] (c) SC [24] (d) Scaling (e) SNS [35] (f) PF[37] (g) Ours

Fig. 19: Comparison of the proposed method in retargeting with some other methods

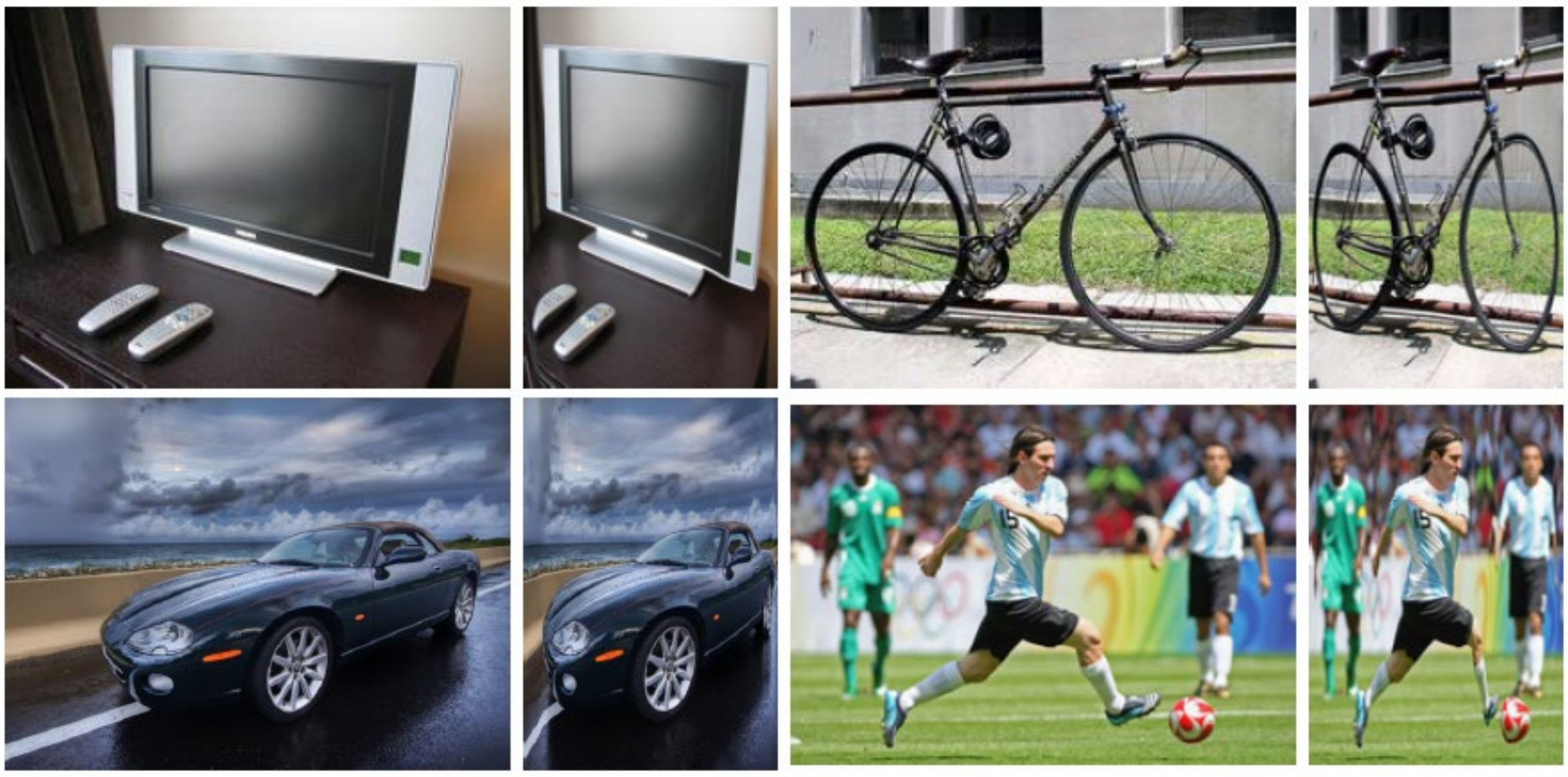

Fig. 20: Some failure cases in image retargeting